\documentclass[runningheads]{llncs}
\usepackage[T1]{fontenc}
\usepackage{graphicx}
\usepackage{amsmath, amssymb} 
\usepackage{booktabs}
\usepackage{multirow}
\usepackage{tabularx}
\usepackage{threeparttable}
\usepackage{float}
\usepackage{enumitem}
\usepackage{xcolor}
\definecolor{darkgreen}{rgb}{0.0,0.5,0.0}
\usepackage{subcaption}
\usepackage{pifont}

\usepackage{tcolorbox}
\usepackage{enumitem}
\tcbuselibrary{skins}
\usepackage[table]{xcolor}
\definecolor{secbg}{HTML}{D6E4F0}
\definecolor{bestbg}{HTML}{D5F5E3}
\usepackage{hyperref}

\begin{document}
\title{Graph-of-Differences: Anatomy-Structured Difference Alignment for Medical Image Re-Identification}
\titlerunning{Graph-of-Differences} 


\author{Nichula Wasalathilaka$^{1,*}$,
Abhijit Das$^{2,*}$,
Imran Razzak$^{2,4}$, \\
Dwarikanath Mahapatra$^{3}$}
\authorrunning{Wasalathilaka et al.}
\institute{$^1$ Khalifa University, Abu Dhabi, UAE.
    $^2$ MBZUAI, Abu Dhabi, UAE. \\
    $^3$ University of Peradeniya, Sri Lanka.
    $^4$ MedOS, Abu Dhabi, UAE.}
    
\maketitle
\let\thefootnote\relax\footnotetext{\textsuperscript{*}Equal contribution. Contact: \textit{dwarikanath.mahapatra@ku.ac.ae}}

\begin{abstract}
Medical image re-identification (MedReID) enables longitudinal patient linkage but remains vulnerable to shortcut learning and produces decisions that clinicians cannot audit against named anatomy. We propose \textbf{Graph-of-Differences (GoD)}, which grounds identity comparisons in explicit anatomical structure. Each image is represented as an anatomy graph whose nodes correspond to named anatomical regions; given an image pair, soft node correspondence is established, and differences are computed over matched anatomy. A graph-level difference alignment objective ties these anatomy-matched differences to the global backbone difference, ensuring the retrieval signal is anchored in homologous structures rather than arbitrary spatial tokens. Explanations are defined over named graph nodes and quantitatively audited via node insertion/deletion tests—replacing unstable pixel heatmaps with verifiable, structure-level evidence. On internal benchmarks, GoD improves Rank-1 by \textbf{+7.1 pp} on fundus and \textbf{+3.1 pp} on CXR over a strong frozen-backbone baseline, with further gains on zero-shot external transfers confirming that anatomy grounding improves both accuracy and generalization. Code released at \href{https://github.com/GenMI-Lab/GoD.git}{https://github.com/GenMI-Lab/GoD.git}.

\keywords{Medical image re-identification \and Anatomy graphs \and Difference alignment \and Auditability }
\end{abstract}


\section{Introduction}
\label{sec:intro}

Medical images carry persistent, patient-specific biometric signals that
survive de-identification established across chest X-rays~\cite{packhaeuser2022biometric,macpherson2023patient},
CT~\cite{heinrich2024ctid}, and histopathology~\cite{ganz2025historeid}.
Medical image re-identification (MedReID) exploits these signals for
longitudinal record linkage, cohort deduplication, and privacy risk
auditing~\cite{tian2025allinone}. As these systems approach clinical
deployment, a fundamental question emerges that accuracy metrics alone cannot
answer: \textit{which anatomical structures is the model actually comparing,
and can a clinician verify this?}

\medskip\noindent\textbf{The gap accuracy cannot close.}
State-of-the-art MedReID (e.g., MaMI~\cite{tian2025allinone}) aligns
inter-image feature differences in token space powerful, but with a structural
limitation: differences are formed without any enforced correspondence to named
anatomical regions. A difference vector may reflect the optic disc in one image
and a vessel junction in the other, or worse, a scanner artefact or exposure
shift~\cite{geirhos2020shortcut}. Post-hoc saliency maps cannot reliably
reveal this: pixel-level heatmaps are unstable under benign perturbations and
unverifiable against named anatomy~\cite{nauta2023explainable}. This is not
an accuracy problem,it is an \emph{auditability} problem where a mislinked record can cause incorrect treatment or a privacy breach.

\begin{figure*}[t]
  \centering
  \setlength{\tabcolsep}{2pt}
  \begin{tabular}{ccc}
    \includegraphics[width=0.2\textwidth]{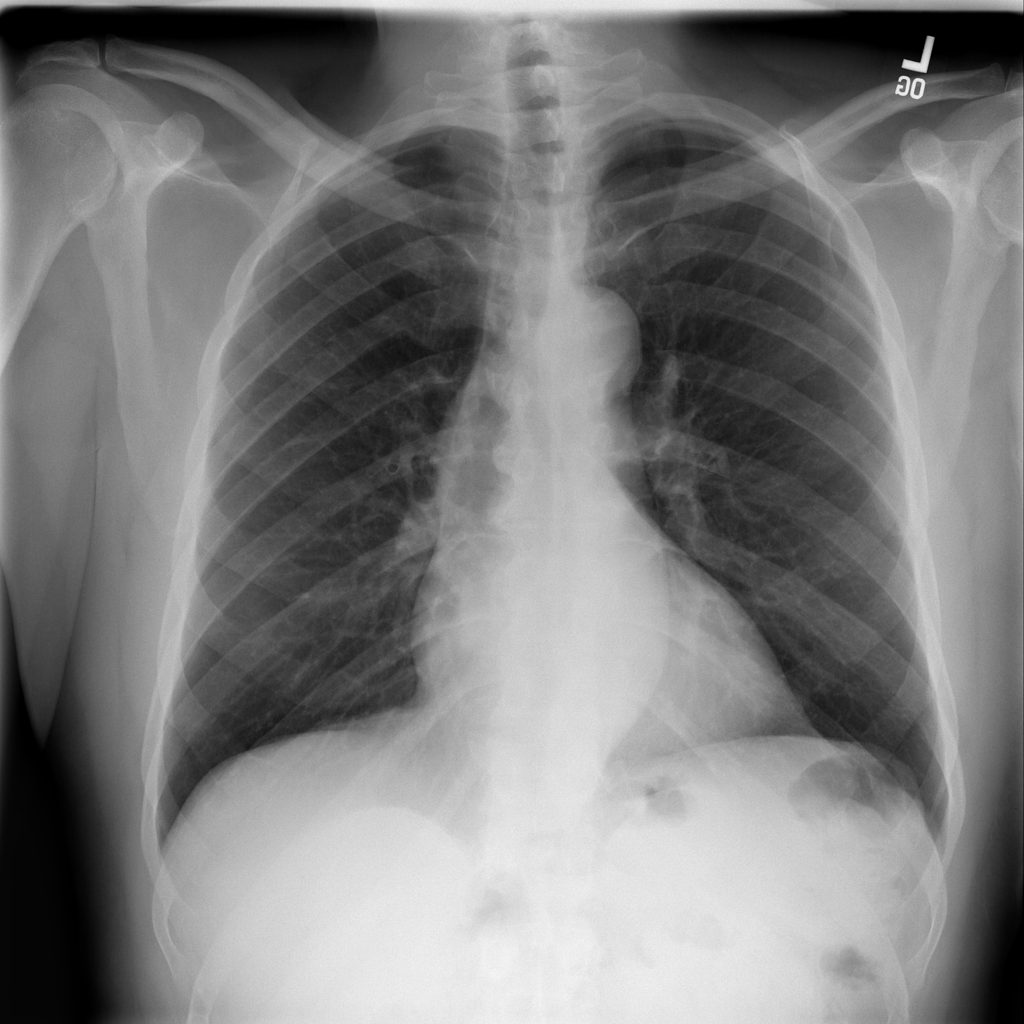} &
    \includegraphics[width=0.2\textwidth]{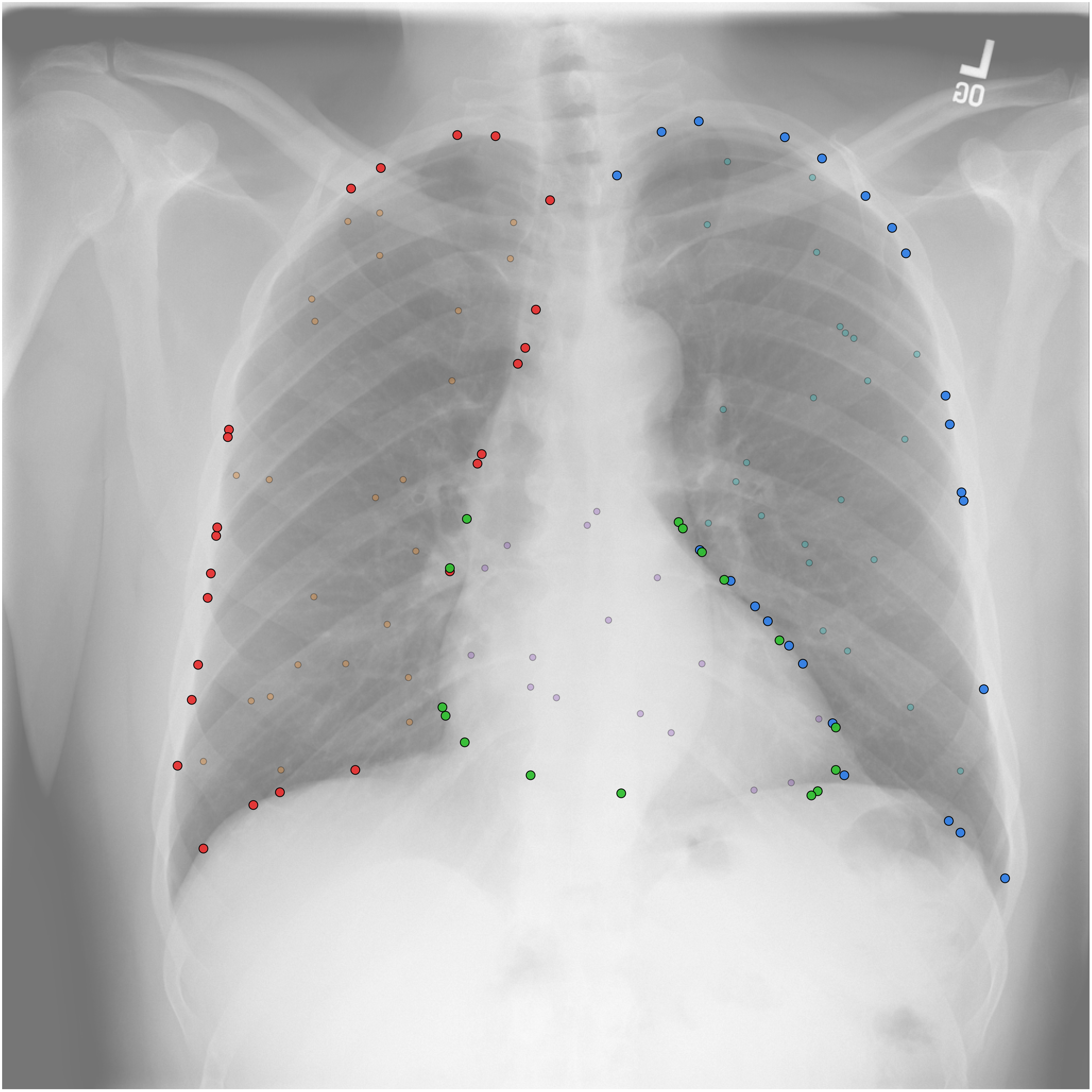} &
    \includegraphics[width=0.2\textwidth]{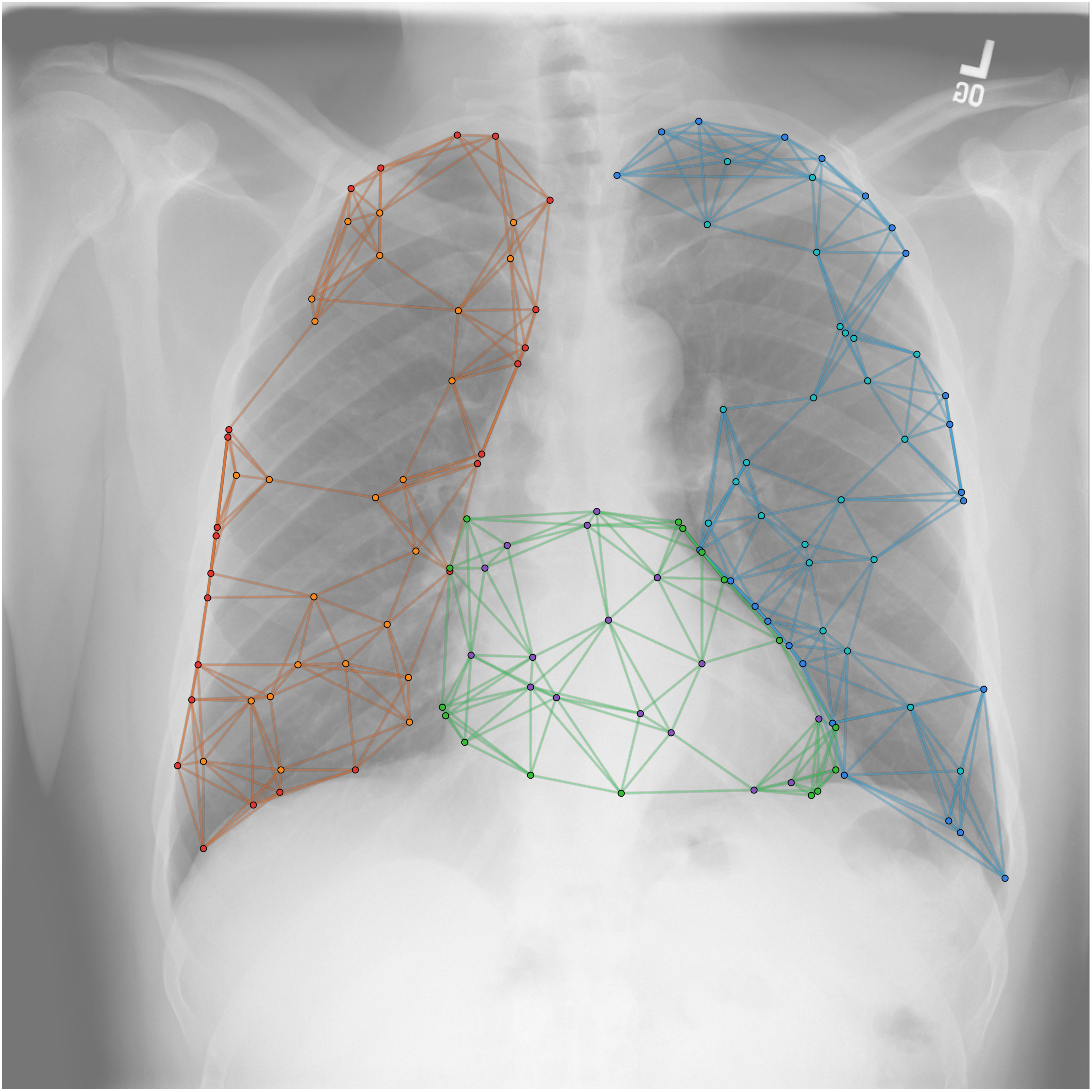}
  \end{tabular}
  \vspace{-4pt}
  \caption{\textbf{CXR anatomy graph construction.} \textbf{(a)}~Input radiograph. \textbf{(b)}~Nodes sampled from CheXmask masks: \textcolor{red}{left lung boundary}, \textcolor{orange}{left interior}, \textcolor{blue}{right lung boundary}, \textcolor{cyan}{right interior}, \textcolor{green}{heart boundary}, \textcolor{violet}{heart interior}. \textbf{(c)}~Graph $G(x)\!=\!(\mathcal{V},\mathcal{E},H)$ with $k$-NN ($k\!=\!6$) and lung-symmetry edges; node features pooled from the frozen map $F(x)$ (Eq.~\ref{eq:node_feat}).}
  \label{fig:pipeline}
\end{figure*}

\noindent\textbf{Our solution.}
We propose \textbf{Graph-of-Differences (GoD)}, which enforces identity
comparisons over \emph{named anatomical structures} the same optic disc, the
same lung boundary, the same cardiac contour-rather than arbitrary spatial
tokens. Each image is represented as an anatomy graph whose nodes are named
anatomical regions constructed by modality-specific parsers. Soft node-to-node correspondence is established across an image pair, differences are computed exclusively over matched anatomy, and a \emph{graph-level difference alignment} objective ties these structured differences to the global backbone signal. Because every comparison is grounded in an identified anatomical node, explanations are attributions over \emph{named structures} with quantitative faithfulness evidence. This per-pair, difference-over-matched-anatomy formulation distinguishes GoD from prior graph-based MedReID such as the Graph Matching Network of Manesco et al.~\cite{manesco2024graph}, which encodes each chest radiograph as a \emph{single-image} graph embedding over anatomical-landmark nodes: it neither establishes per-pair correspondence between named structures, nor computes differences over matched nodes, nor audits node-level attributions for faithfulness. Our contributions are-

\begin{enumerate}[leftmargin=*,itemsep=2pt,topsep=2pt,parsep=0pt]
  \item \textbf{Anatomy-structured difference alignment.} A graph-level objective computes differences over soft-corresponded anatomical nodes and aligns them with the global difference signal, grounding identity decisions in homologous anatomy and mitigating shortcut learning.
  \item \textbf{Structure-level interpretability with faithfulness evidence.} Node-level attributions over named anatomical structures are quantitatively audited via insertion/deletion curves, replacing unverifiable pixel heatmaps; full clinical auditability further requires radiologist user studies (future work).
  \item \textbf{Consistent gains across benchmarks.} On fundus (ODIR, Messidor-2) and CXR (NIH14, CheXpert), GoD improves Rank-1 by \textbf{+7.1 pp} and \textbf{+3.1 pp} over MaMI internally, with largest relative gains on zero-shot external splits.
\end{enumerate}

\begin{tcolorbox}[
  colback=green!10,
  colframe=green!40!black,
  leftrule=2pt, rightrule=0.5pt, toprule=0.5pt, bottomrule=0.5pt,
  boxsep=3pt, left=2pt, right=2pt, top=2pt, bottom=2pt,
  title={\scriptsize\bfseries\textcolor{white}{SOTA methods vs.\ Graph-of-Differences (GoD)}},
  fonttitle=\scriptsize\bfseries,
  coltitle=white,
  attach boxed title to top left={yshift=-2mm, xshift=4mm},
  boxed title style={colback=green!40!black, colframe=green!40!black,
    boxrule=0pt, arc=1pt}]

\scriptsize
\textbf{SOTA Methods like MaMI} asks: \textit{Can a single model re-identify patients across
diverse modalities?} It achieves this via token-space difference alignment
with a modality-adaptive backbone (ComPA), but without enforced anatomical
correspondence—leaving identity decisions unverifiable against named anatomy.

\smallskip
\textbf{GoD} asks: \textit{Can we guarantee that identity decisions are
grounded in homologous anatomy, and can a clinician verify this?} This
requires an architectural commitment—an explicit anatomy graph, enforced
node correspondence, and a structured difference objective—that cannot be
retrofitted post-hoc. GoD uses MaMI's encoder as a \emph{frozen} prior,
so all gains arise purely from anatomy-structured difference alignment.
\end{tcolorbox}


\section{Method}
\label{sec:method}

\textbf{Task.} Given query $x_a$ and a gallery, MedReID retrieves same-patient
images by ranking via cosine similarity $s(x_a,x_b)=\cos(z(x_a),z(x_b))$,
where $z(\cdot)$ is a learned retrieval embedding~\cite{macpherson2023patient,packhaeuser2022biometric,tian2025allinone}.

\noindent\textbf{Backbone.} We adopt the pretrained MaMI encoder~\cite{tian2025allinone}
(ViT-B\cite{Dosovitskiy2021ViT} + ComPA) as a \emph{frozen} Siamese feature extractor, producing a
global descriptor $g(x)\!\in\!\mathbb{R}^{768}$ and spatial feature map
$F(x)\!\in\!\mathbb{R}^{768\times14\times14}$. Freezing isolates the
contribution of anatomy-structured difference alignment from backbone
fine-tuning; only the graph encoder, fusion module, and projection head
$\phi(\cdot)$ are trained.

\noindent\textbf{GoD pipeline (Fig.~\ref{fig:pipeline}).}
For each image: (i)~extract backbone features, (ii)~build an anatomy graph with
backbone-pooled node features, (iii)~fuse graph descriptor $g_G(x)$ with
$g(x)$ to form $z(x)$. Per pair: establish soft node correspondence, compute
node-wise differences over matched anatomy, and align the graph difference with
the backbone difference via $\mathcal{L}_{\mathrm{diff}}$  grounding identity
decisions in homologous structures and enabling node-level auditing.

\subsection{Anatomy Graph Construction}
\label{sec:graph_construction}

Each image is represented as $G(x)\!=\!(\mathcal{V},\mathcal{E},H)$: nodes
$\mathcal{V}$ are named anatomical regions, edges $\mathcal{E}$ sparse
geometric relations, and features $H\!=\!\{h_i\}$ are masked-average-pooled
from $F(x)$, inheriting pretrained semantics while remaining spatially
localised:
\begin{equation}
  h_i = \frac{\sum_{u,v}M_i(u,v)\,F(x)_{u,v}}{\sum_{u,v}M_i(u,v)+\epsilon}.
  \label{eq:node_feat}
\end{equation}

For \textbf{fundus} ($N\!\le\!34$), nodes cover optic-disc/macula landmarks,
vessel junctions, and polar anchor nodes, canonicalised to a nasal-side optic
disc orientation. For \textbf{CXR} ($N\!=\!128$), CheXmask~\cite{Gaggion2025CheXmask}
parses left lung, right lung, and heart masks; we sample 24 boundary $+$ 24
interior nodes per lung and $16\!+\!16$ for the heart, with farthest-point
sampling\cite{Qi2017PointNetPP} ensuring uniform spatial coverage. Each node connects to its
$k\!=\!6$ nearest neighbours\cite{Wang2018DGCNN} (plus self-loops); CXR graphs additionally
include left--right lung symmetry edges.

\begin{figure}[t]
  \centering
  \includegraphics[width=0.8\textwidth]{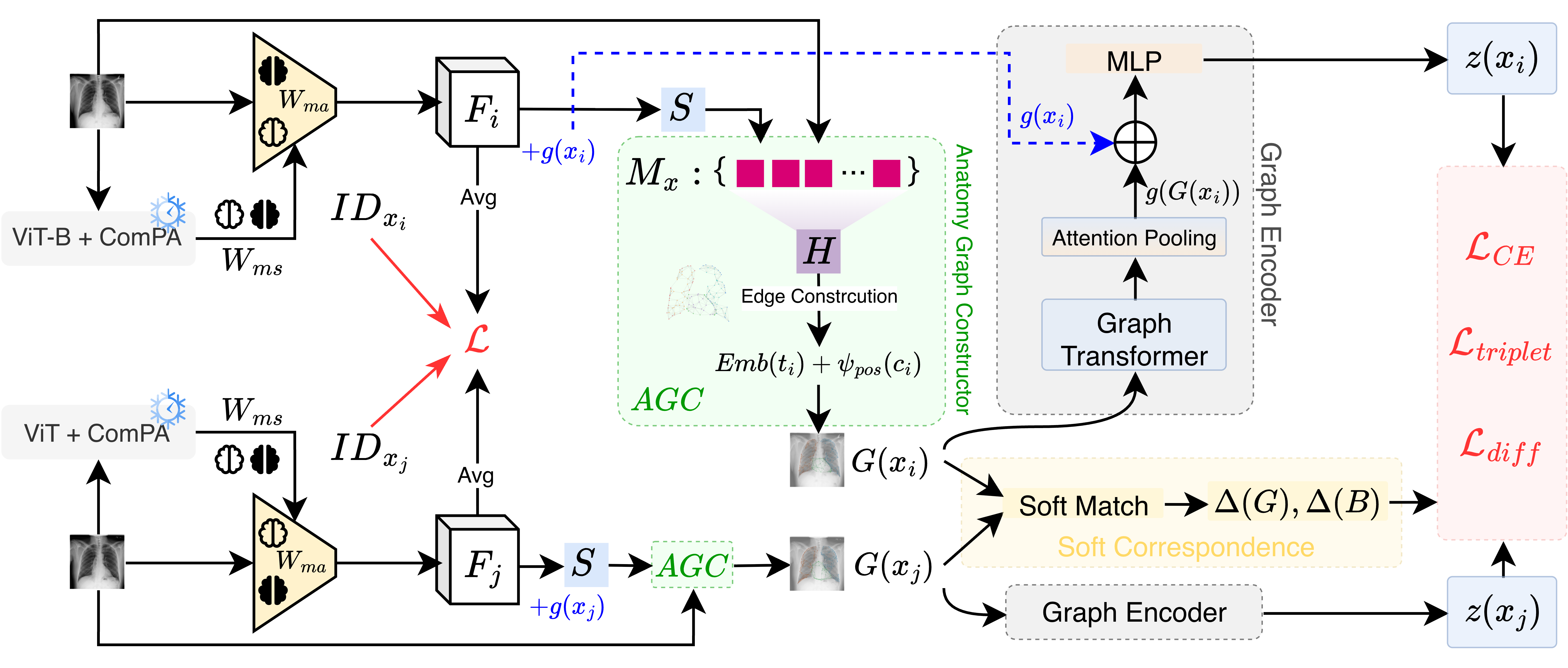}
    \caption{\textbf{Overview of Graph-of-Differences (GoD).} A frozen MaMI encoder (ViT-B + ComPA) yields a global descriptor $g(x)$ and feature map $F(x)$. The Anatomy Graph Constructor~(AGC) pools node features from $F(x)$ (Eq.~\ref{eq:node_feat}) and builds graphs with $k$-NN and symmetry edges. A shared Graph Encoder yields $g_G(x)$, fused with $g(x)$ into the embedding $z(x)$. Per pair, soft correspondence computes anatomy-matched differences $\Delta_G$, aligned with $\Delta_B$ via $\mathcal{L}_{\mathrm{diff}}$ (Eq.~\ref{eq:ldiff}).}
\label{fig:architecture}
\end{figure}

\subsection{Graph Encoder and Retrieval Embedding}
\label{sec:graph_encoder}

\textbf{Graph transformer.} A lightweight masked graph transformer\cite{Vaswani2017Attention,Velickovic2018GAT}
($L\!=\!2$, $H_{\mathrm{att}}\!=\!4$) encodes each node as:
\begin{equation}
  \tilde{h}_i = h_i + \mathrm{Emb}(t_i) + \phi_{\mathrm{pos}}(c_i),
\end{equation}
combining appearance, learnable type embedding, and coordinate MLP. Attention
is masked to $k$-NN neighbourhoods\cite{Wang2018DGCNN}, enforcing anatomically local message
passing and yielding contextualised embeddings $H'\!=\!\{h_i'\}_{i=1}^N$.

\noindent\textbf{Pooling and fusion.} A learnable attention vector $p$ pools $H'$ into
a graph descriptor\cite{Ilse2018AttnMIL} $g_G(x)\!\in\!\mathbb{R}^{768}$:
\begin{equation}
  \alpha_i=\mathrm{softmax}\!\left(\tfrac{p^\top h_i'}{\sqrt{d}}\right),
  \quad g_G(x)=\textstyle\sum_i \alpha_i h_i'.
\end{equation}
$g(x)$ and $g_G(x)$ carry complementary holistic and anatomy-structured
signals; we concatenate and project to the final $\ell_2$-normalised embedding:
\begin{equation}
  z(x)=\phi\!\left([g(x)\!\parallel\!g_G(x)]\right)\big/\!\big\|{\cdot}\big\|_2,
  \quad [g\!\parallel\!g_G]\!\in\!\mathbb{R}^{1536},\; z(x)\!\in\!\mathbb{R}^{512}.
\end{equation}

\subsection{Graph-of-Differences and Training}
\label{sec:diff_alignment}

\textbf{Soft correspondence.} Given node embeddings
$H_a\!\in\!\mathbb{R}^{N_a\times D}$, $H_b\!\in\!\mathbb{R}^{N_b\times D}$
for a pair $(x_a,x_b)$, we match nodes via temperature-scaled similarity and
row-wise softmax, yielding matched embeddings $\tilde{H}_b\!=\!PH_b$:
\begin{equation}
  S=\tfrac{1}{\tau}\,\mathrm{norm}(H_a)\mathrm{norm}(H_b)^\top,\qquad
  P_{ij}=\mathrm{softmax}_j(S_{ij}).
\end{equation}
Unmatched nodes receive low correspondence mass, making this robust to parser
imperfection and unequal node counts.

\noindent\textbf{Difference alignment.} Node-wise differences over matched anatomy and
their mean-pooled aggregate are:
\begin{equation}
  \delta_i=h_i(x_a)-\tilde{h}_i(x_b),\qquad
  \Delta_G=\tfrac{1}{N_a}\textstyle\sum_i\delta_i.
\end{equation}
We align $\Delta_G$ with the global backbone difference
$\Delta_B\!=\!g(x_a)\!-\!g(x_b)$ via a learnable projection $W$:
\begin{equation}
  \mathcal{L}_{\mathrm{diff}}=1-\cos(W\Delta_G,\,\Delta_B),
  \label{eq:ldiff}
\end{equation}
enforcing that anatomy-matched differences explain the global retrieval signal. Because $\mathcal{L}_{\mathrm{diff}}$ operates directly on $\{\delta_i\}$, the insertion/deletion audit targets exactly the component the attributions describe; propagating attributions through the fusion into $z(x)$ would require learned credit assignment and is deferred (Sec.~\ref{sec:discussion}) rather than approximated. The full objective is\cite{Hermans2017TripletLoss}:
\begin{equation}
  \mathcal{L}=\mathcal{L}_{\mathrm{CE}}+\mathcal{L}_{\mathrm{triplet}}
  +\lambda\,\mathcal{L}_{\mathrm{diff}},\quad\lambda\!=\!0.1,\;m\!=\!0.3.
\label{eq:loss}
\end{equation}


\section{Experiments}
\label{sec:experiments}

\noindent\textbf{Datasets and Evaluation.}We evaluate on fundus and CXR under \textbf{internal} (held-out split) and \textbf{external} (zero-shot transfer) protocols with patient-disjoint splits, reporting cumulative-matching-characteristic \textbf{Rank-1 retrieval accuracy (R1)}, i.e.\ the fraction of queries whose top-ranked gallery image is a true same-patient match, and \textbf{mean Average Precision (mAP)}. External gains cannot be attributed to overfitting and directly test whether anatomy grounding improves generalisation under real-world acquisition variability.

\begin{table}[t]
\scriptsize
\centering
\caption{Evaluation data. Multi-site and cross-dataset settings
stress-test robustness to acquisition variability where shortcut-exploiting
models degrade most.}
\label{tab:datasets}
\setlength{\tabcolsep}{4pt}
\begin{tabular}{llllrr}
\toprule
\textbf{Modality} & \textbf{Split} & \textbf{Dataset}
& \textbf{Source} & \textbf{Patients} & \textbf{Images} \\
\midrule
\multirow{2}{*}{Fundus}
  & Internal & ODIR-2019~\cite{ODIR2019}
  & Multi-site, multi-device &  5,000 &  10,000 \\
  & External & Messidor-2\cite{Decenciere2014MESSIDOR}
  & Macula-centred, paired   &    874 &   1,748 \\
\midrule
\multirow{2}{*}{CXR}
  & Internal & NIH ChestX-ray14\cite{Wang2017ChestXray14}
  & 14-label, NLP-mined      & 30,805 & 112,120 \\
  & External & CheXpert\cite{Irvin2019CheXpert}
  & Frontal + lateral        & 65,240 & 224,316 \\
\bottomrule
\end{tabular}
\end{table}

\noindent\textbf{Implementation Details.} The MaMI backbone (ViT-B + ComPA) is \emph{frozen} throughout; only the graph encoder, fusion module, projection head, and identity classifier are trained via AdamW with cosine annealing over 30 epochs. All hyperparameters are reported in Table~\ref{tab:impl_ablations}. Pairs are mined with a BatchHard sampler (batch size 64); inputs are resized to $224\times224$ with ImageNet normalisation and label smoothing $0.1$. For external evaluation, models trained on the internal split are applied zero-shot to the external dataset with identical preprocessing and no fine-tuning. All experiments are run on 2 NVIDIA Quadro GV100 GPUs (32 GB each), Intel Xeon Gold 5218R CPUs (2 sockets, 80 cores, 80 threads), running Ubuntu 22.04.5 LTS.


\section{Results}
\label{sec:results}

\noindent\textbf{Re-Identification Performance.} Table~\ref{tab:main_results_modalities} reports Rank-1 and mAP across all four benchmarks. GoD outperforms the frozen-backbone baseline on every split: \textbf{+3.1 pp R1 / +4.3 pp mAP} on CXR and \textbf{+7.1 pp R1 / +2.6 pp mAP} on fundus (internal), with gains arising entirely from the graph encoder, fusion module, and $\mathcal{L}_{\mathrm{diff}}$. The gain is orthogonal to backbone tuning: on a fine-tuned MaMI encoder GoD still adds \textbf{+2.5 pp} (CXR) and \textbf{+5.4 pp} (fundus) R1 (Table~\ref{tab:finetune}), with the best absolute result on the fine-tuned backbone. Adding a graph branch \emph{without} correspondence or alignment yields marginal or negative changes, confirming gains come from \emph{aligning differences over matched anatomy}, not graph structure alone. Gains hold on zero-shot external transfers (\textbf{+2.8 pp} CXR, \textbf{+0.8 pp} fundus R1): on Messidor-2, GoD reaches R1 $=0.6579$ (95\% CI $[0.6491,0.6664]$) versus the backbone at $0.6499$ ($[0.6398,0.6604]$), indicating identity-stable anatomical signals rather than dataset-specific shortcuts.

\begin{table}[t]
\centering
\tiny
\caption{MedReID performance across modalities.}
\label{tab:main_results_modalities}
\begin{threeparttable}
\setlength{\tabcolsep}{1pt}
\begin{tabular}{lcc|cc|cc|cc}
\toprule
& \multicolumn{4}{c}{CXR} & \multicolumn{4}{c}{Fundus} \\
\cmidrule(lr){2-5}\cmidrule(lr){6-9}
& \multicolumn{2}{c}{Internal (NIH14)} & \multicolumn{2}{c}{External (CheXpert)} &
\multicolumn{2}{c}{Internal (ODIR)} & \multicolumn{2}{c}{External (Messidor-2)} \\
Method & R1$\uparrow$ & mAP$\uparrow$ & R1$\uparrow$ & mAP$\uparrow$ &
R1$\uparrow$ & mAP$\uparrow$ & R1$\uparrow$ & mAP$\uparrow$ \\
\midrule
Backbone-only
& 0.8723 & 0.7867 & 0.3084 & 0.3788
& 0.6396 & 0.3631 & 0.6499 & 0.7432 \\
Backbone + Graph (w/o $\mathcal{L}_{diff}$)
& 0.8659 & 0.7686 & 0.3068 & 0.3710
& 0.6357 & 0.3610 & 0.6327 & 0.7336 \\
Full (Graph-of-Differences)
& 0.9033 & 0.8294 & 0.3359 & 0.3926
& \textbf{0.7111} & \textbf{0.3894} & 0.6579 & 0.7492 \\
\bottomrule
\end{tabular}

\end{threeparttable}
\end{table}

\begin{table}[t]
\centering
\tiny
\caption{\textbf{Component ablation.} \textbf{(A)}~backbone-only;
\textbf{(B)}~+graph; \textbf{(C)}~+correspondence;
\textbf{(D)}~full GoD. Mean\,$\pm$\,std over three seeds.}
\label{tab:component_ablation}
\setlength{\tabcolsep}{4pt}
\renewcommand{\arraystretch}{1.05}
\begin{tabular}{lccc cc cc}
\toprule
& \multicolumn{3}{c}{Components}
& \multicolumn{2}{c}{Fundus (ODIR)}
& \multicolumn{2}{c}{CXR (NIH14)} \\
\cmidrule(lr){2-4}\cmidrule(lr){5-6}\cmidrule(lr){7-8}
& Graph & Corr. & $\mathcal{L}_{\mathrm{diff}}$
& R1$\uparrow$ & mAP$\uparrow$
& R1$\uparrow$ & mAP$\uparrow$ \\
\midrule
(A) & -- & -- & --
    & 0.6396\,$\pm$\,0.0098 & 0.3631\,$\pm$\,0.0032
    & 0.8723\,$\pm$\,0.0005 & 0.7867\,$\pm$\,0.0007 \\
(B) & \checkmark & -- & --
    & 0.6357\,$\pm$\,0.0020 & 0.3610\,$\pm$\,0.0018
    & 0.8659\,$\pm$\,0.0041 & 0.7686\,$\pm$\,0.0074 \\
(C) & \checkmark & \checkmark & --
    & 0.6569\,$\pm$\,0.0096 & 0.3652\,$\pm$\,0.0119
    & 0.8797\,$\pm$\,0.0068 & 0.7908\,$\pm$\,0.0054 \\
(D) & \checkmark & \checkmark & \checkmark
    & \textbf{0.7111\,$\pm$\,0.0013} & \textbf{0.3894\,$\pm$\,0.0010}
    & \textbf{0.9033\,$\pm$\,0.0066} & \textbf{0.8294\,$\pm$\,0.0113} \\
\bottomrule
\end{tabular}
\end{table}

\begin{table}[t]
\centering
\tiny
\begin{minipage}[t]{0.48\textwidth}
\centering
\caption{\textbf{GoD is orthogonal to backbone fine-tuning} (R1$\uparrow$). GoD improves both frozen and fine-tuned MaMI; best combines both. Frozen: three-seed means (Table~\ref{tab:component_ablation}).}
\label{tab:finetune}
\setlength{\tabcolsep}{8pt}
\begin{tabular}{lcc}
\toprule
Configuration & CXR & Fundus \\
\midrule
Backbone, frozen & 0.8723 & 0.6396 \\
Backbone, fine-tuned & 0.8841 & 0.6712 \\
GoD, frozen & 0.9033 & 0.7111 \\
GoD, fine-tuned & \textbf{0.9089} & \textbf{0.7253} \\
\bottomrule
\end{tabular}
\end{minipage}
\hfill
\begin{minipage}[t]{0.48\textwidth}
\centering
\caption{\textbf{Parser-robustness stress test} (R1$\uparrow$). GoD stays above the no-graph baseline through 50\% node dropout and 10-px dilation.}
\label{tab:robustness}
\setlength{\tabcolsep}{8pt}
\begin{tabular}{lcc}
\toprule
Perturbation & CXR & Fundus \\
\midrule
Clean (default) & 0.9033 & 0.7111 \\
10\% nodes dropped & 0.8987 & 0.7044 \\
30\% nodes dropped & 0.8821 & 0.6852 \\
50\% nodes dropped & 0.8543 & 0.6517 \\
5-px dilation & 0.8951 & 0.7038 \\
10-px dilation & 0.8782 & 0.6859 \\
No graph (frozen) & 0.8723 & 0.6396 \\
\bottomrule
\end{tabular}
\end{minipage}
\end{table}

\subsection{Ablation Study}
\label{sec:ablation}

Table~\ref{tab:component_ablation} progressively adds components over three
seeds. A graph branch alone (B) slightly hurts performance; adding soft
correspondence (C) recovers and surpasses backbone-only; the full model with
$\mathcal{L}_{\mathrm{diff}}$ (D) yields the largest gains (\textbf{+7.1 pp}
fundus, \textbf{+3.1 pp} CXR R1). The monotonic improvement validates that
difference alignment—not graph expressivity—is the key driver.

\subsection{Explainability and Auditability}
\label{sec:xai_results}

\begin{figure*}[t]
  \centering
  \begin{subfigure}[t]{0.49\textwidth}
    \centering
    \includegraphics[width=\textwidth]{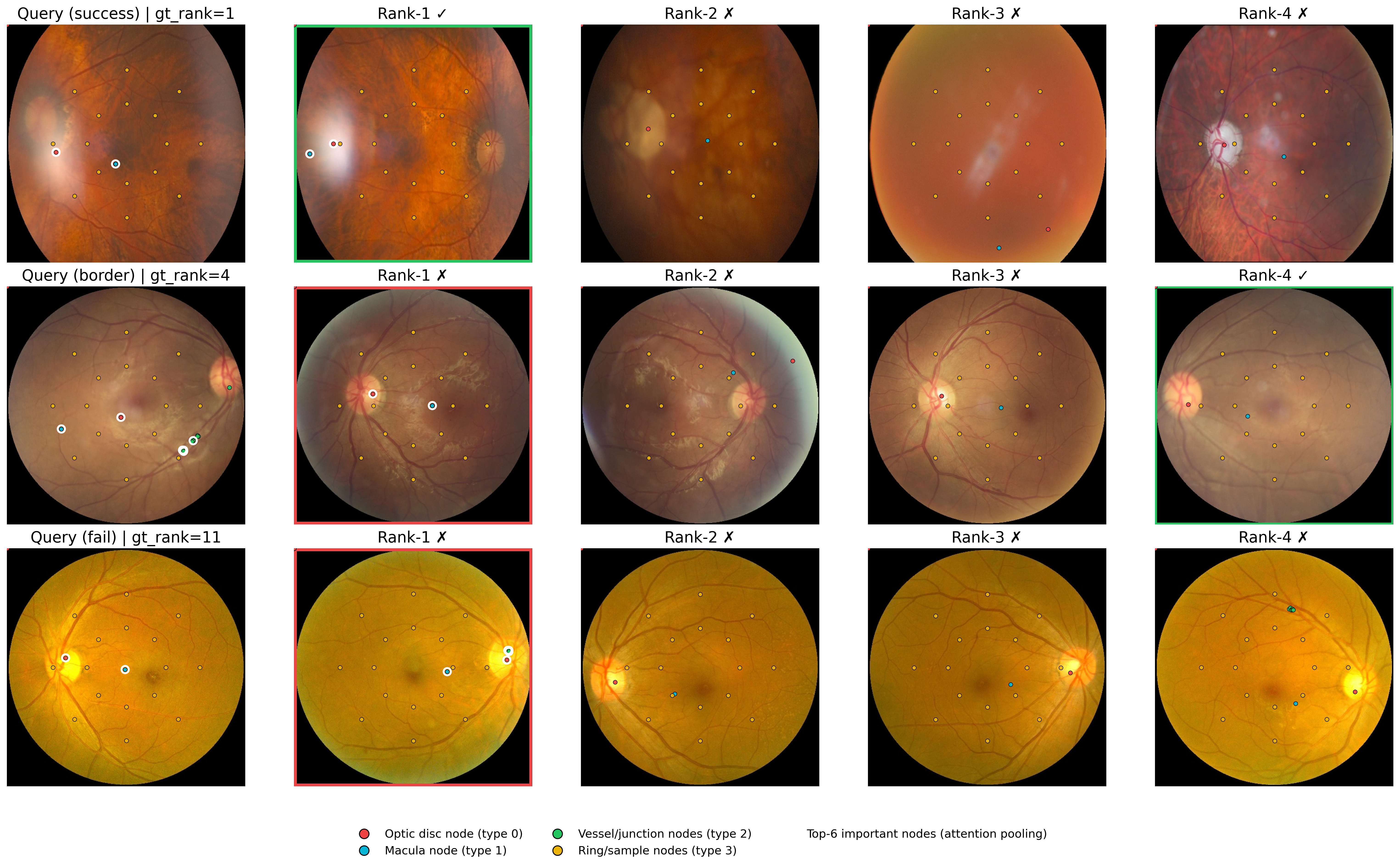}
    \caption{Fundus (ODIR). Nodes: optic disc, macula, vessel junction,
    ring/anchor. Highlighted = top-attended nodes in $g_G$.}
    \label{fig:xai_fundus_cases}
  \end{subfigure}
  \hfill
  \begin{subfigure}[t]{0.49\textwidth}
    \centering
    \includegraphics[width=\textwidth]{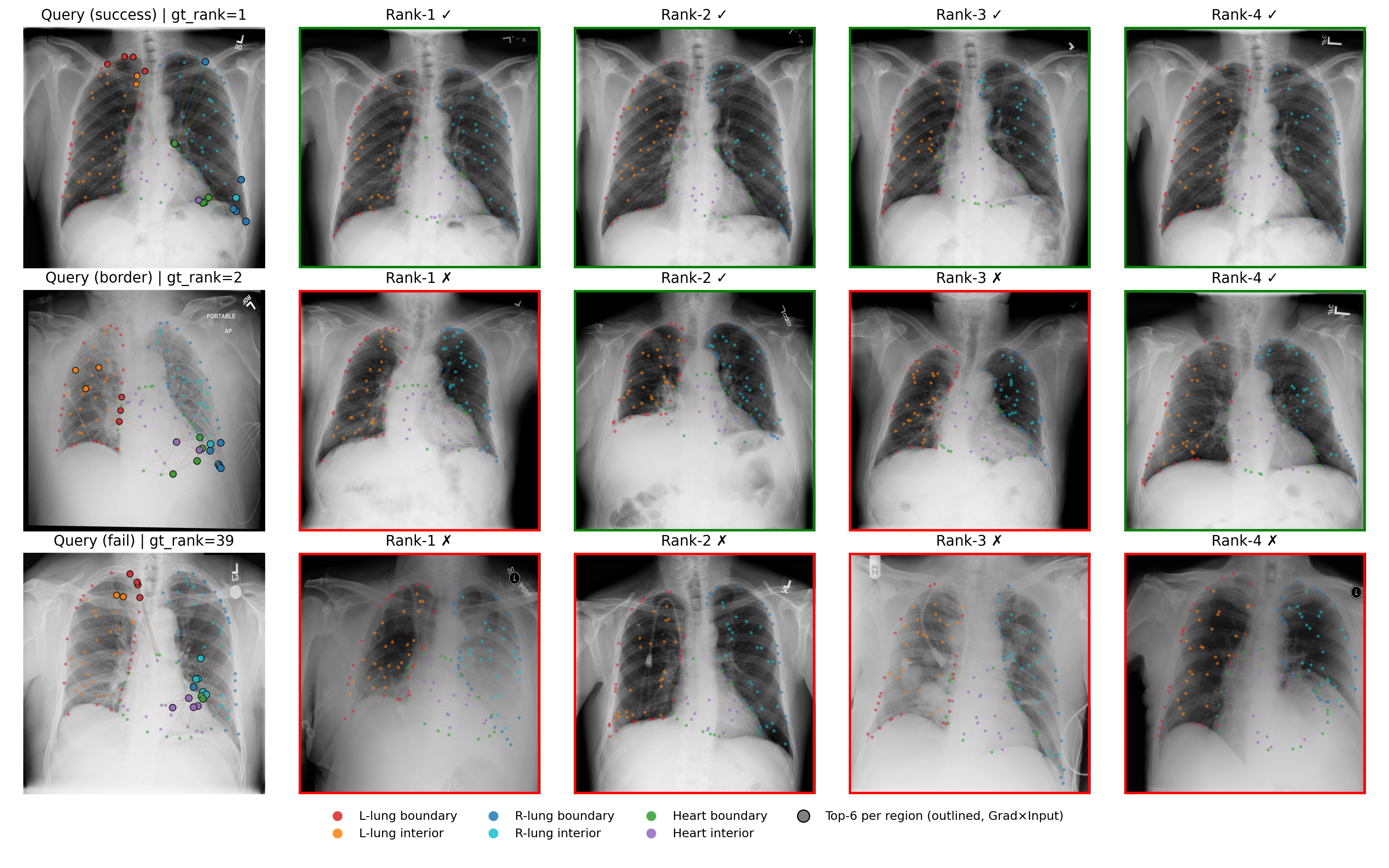}
    \caption{CXR (NIH14). Nodes: left/right lung and heart
    boundary/interior (CheXmask). Highlighted = top-$k$
    Grad$\!\times\!$Input on $g_G$.}
    \label{fig:xai_cxr_cases}
  \end{subfigure}
  \caption{\textbf{Anatomy-aware retrieval explanations.} Each row: one query and top-4 retrievals; \textcolor{green}{green}\,/\,\textcolor{red}{red} borders = correct\,/\,incorrect. Successful retrievals concentrate attribution on stable structures; failures are diffuse.}
\label{fig:xai_combined}
\end{figure*}

\textbf{Qualitative explanations (Fig.~\ref{fig:xai_combined}).}
In successful retrievals, high-attribution nodes concentrate on anatomically
stable structures—optic disc and vessel junctions (fundus); lung boundaries
and cardiac contours (CXR). Borderline and failure cases show diffuse or
peripheral attributions, consistent with confounders such as exposure shifts
or device artefacts. Because attributions are over \emph{named nodes}, a
clinician can directly interrogate which structures drove the decision—
something pixel heatmaps cannot support.

\begin{figure}[t]
  \centering
  \includegraphics[width=0.48\textwidth]{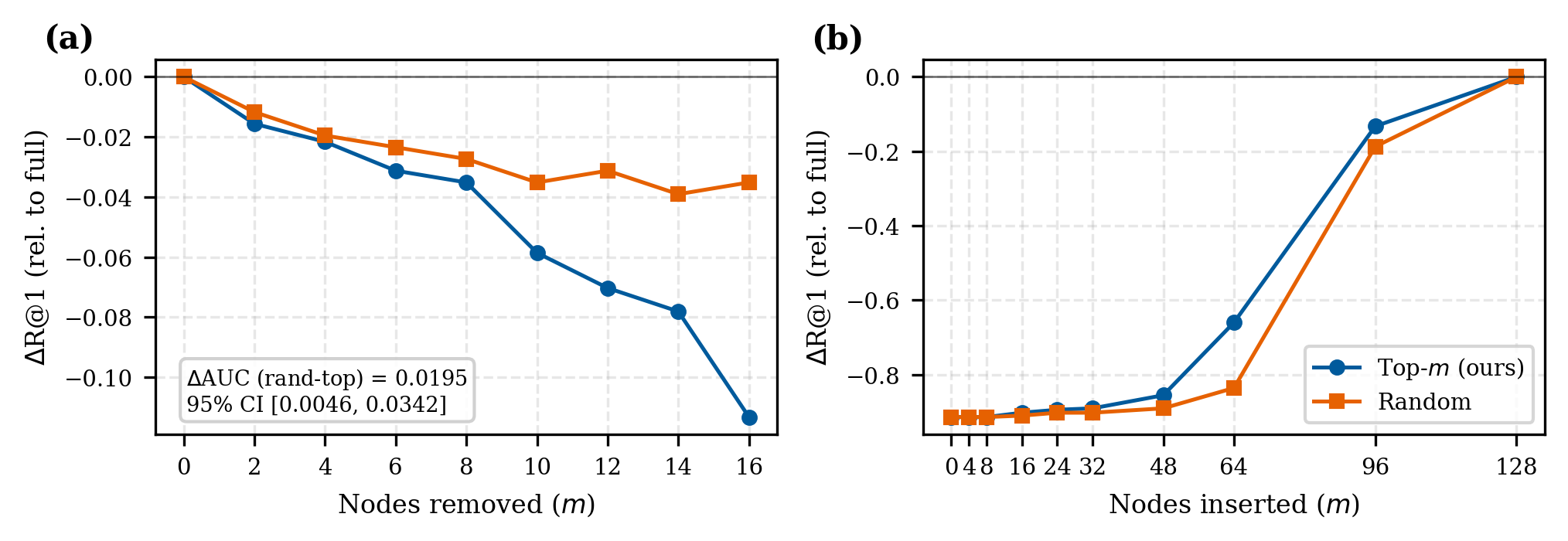}\hfill
  \includegraphics[width=0.48\textwidth]{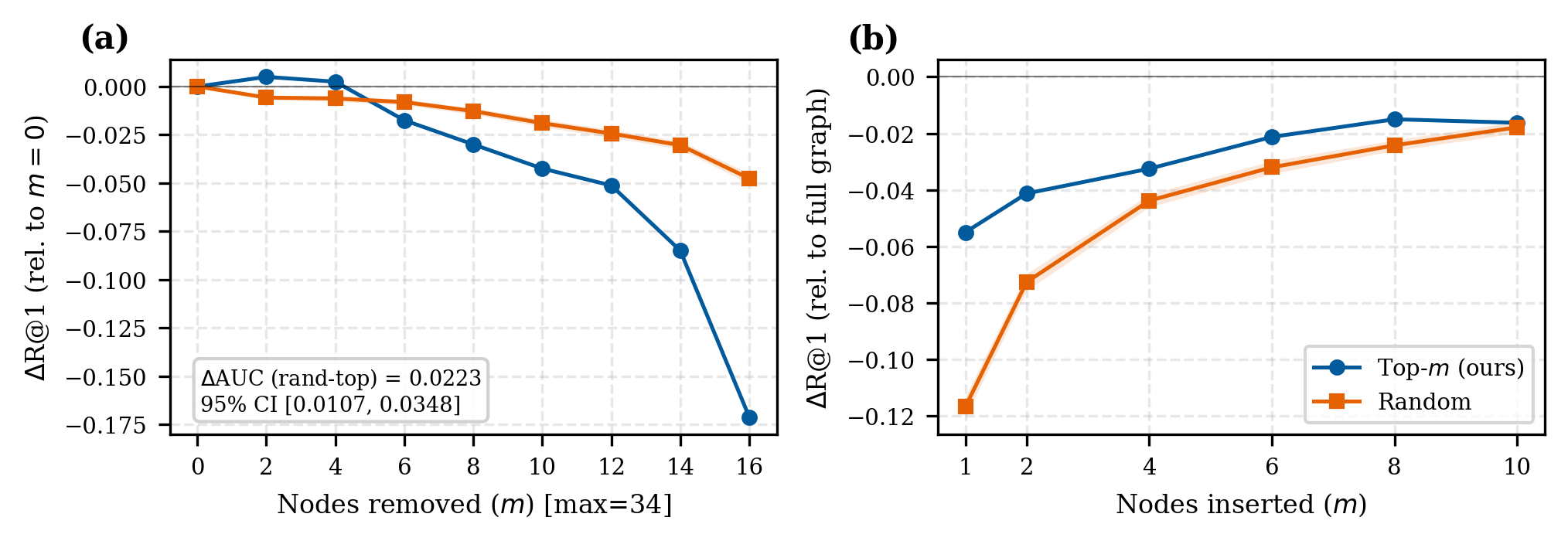}
    \caption{\textbf{Node faithfulness audit.} Removing top-$m$ attributed nodes degrades R1 more than random deletion (CXR: $\Delta\mathrm{AUC}_{\mathrm{R1}}\!=\!0.0195$, CI $[0.0046,\,0.0342]$; Fundus: $0.0223$, CI $[0.0107,\,0.0348]$); insertion restores R1 faster at small $m$. This confirms attributed nodes are causally relied upon, not spurious gradients.}
\label{fig:xai_deletion_insertion}
\end{figure}

\noindent\textbf{Quantitative faithfulness audit (Fig.~\ref{fig:xai_deletion_insertion}, Table~\ref{tab:impl_ablations}).}
Counterfactual node deletion degrades R1 more than random removal
($\Delta\mathrm{AUC}_{\mathrm{R1}}\!=\!0.0195$, CI [0.0046, 0.0342] on
NIH14; $0.0223$, CI $[0.0107,0.0348]$ on ODIR), and insertion restores
performance faster than random at small $m$. This confirms attributions
identify nodes the model \emph{genuinely relies on}—not spurious gradients—
distinguishing GoD from post-hoc saliency~\cite{nauta2023explainable}.
Graph granularity peaks at $1.5\times$ node scaling; sparser graphs lose
discriminative structure, denser graphs dilute correspondence, validating
our default sizes ($N\!\le\!34$ fundus; $N\!=\!128$ CXR).

        \begin{table}[t]
        \centering
        \caption{\textbf{Implementation details and ablations.} Node faithfulness uses Grad$\times$Input (mean, 95\,\% bootstrap CI; higher is better). Graph granularity scales node count vs.\ default ($1.0\times$).}
        \label{tab:impl_ablations}
        \setlength{\tabcolsep}{4.5pt}
        \renewcommand{\arraystretch}{1}
        \resizebox{\textwidth}{!}{%
        \tiny
        \begin{tabular}{l l l}
        \toprule
        \rowcolor{secbg}
        \multicolumn{3}{l}{\textbf{Optimisation}} \\
        Optimizer : AdamW (wd $10^{-4}$) &
        LR backbone : $10^{-3}\!\to\!10^{-5}$ cos. &
        Epochs : 30 \\
        LR GoD : $5\!\times\!10^{-4}\!\to\!10^{-5}$ cos. &
        & \\
        \midrule
        \rowcolor{secbg}
        \multicolumn{3}{l}{\textbf{Losses \& Graph}} \\
        $\lambda_{\mathrm{tri}},\,\lambda_{\mathrm{diff}}$ : 1.0, 0.1 &
        Label smooth.\ $\epsilon$ : 0.1 &
        Margin $m$, scale $s$ : 0.3, 30 \\
        $\tau, k, L, H_{\mathrm{att}}$ : 0.1, 6, 2, 4 &
        \multicolumn{2}{l}{Nodes : Fundus ${\le}34$ / CXR 128}\\
        \bottomrule
        \end{tabular}%
        }%
        
        \vspace{1pt}
        \setlength{\tabcolsep}{2pt}
        \resizebox{\textwidth}{!}{%
        \tiny
        \begin{tabular}{lc @{\hskip 10pt} l cc cc}
        \toprule
        \rowcolor{secbg}
        \multicolumn{2}{l}{\textbf{Node Faithfulness} ($\Delta$Del-AUC$_{\mathrm{R1}}\!\uparrow$)}
        & \multicolumn{5}{l}{\cellcolor{secbg}\textbf{Graph Granularity}} \\
        \cmidrule(r){1-2}\cmidrule(l){3-7}
        & &
        & \multicolumn{2}{c}{\textit{CXR (NIH14)}}
        & \multicolumn{2}{c}{\textit{Fundus (ODIR)}} \\
        \cmidrule(lr){4-5}\cmidrule(l){6-7}
        \textit{Dataset} & \textit{Score [CI]}
        & \textit{Ratio} & R1 & mAP & R1 & mAP \\
        \cmidrule(r){1-2}\cmidrule(l){3-7}
        ODIR (Fundus) & $0.0223\;[.011,\,.035]$
        & $0.5\times$ & 0.9015 & 0.8297 & 0.6910 & 0.3842 \\
        NIH14 (CXR) & $0.0195\;[.005,\,.034]$
        & $0.75\times$ & 0.9064 & 0.8316 & 0.6983 & 0.3865 \\
        & & $1.0\times$ & 0.9105 & 0.8423 & 0.7124 & 0.3904 \\
        & & \cellcolor{bestbg}$\mathbf{1.5\times}$ & \cellcolor{bestbg}\textbf{0.9128} & \cellcolor{bestbg}\textbf{0.8438}
        & \cellcolor{bestbg}\textbf{0.7159} & \cellcolor{bestbg}\textbf{0.3911} \\
        & & $2.0\times$ & 0.9037 & 0.8314 & 0.7118 & 0.3902 \\
        \bottomrule
        \end{tabular}%
        }%
        \end{table}
        
\section{Discussion}
\label{sec:discussion}
The ablation results establish that anatomy grounding only pays off when correspondence and difference alignment act together—a graph branch alone is inert, and gains emerge specifically from comparing \emph{matched} nodes. That the largest improvements appear on external, zero-shot benchmarks (CheXpert, Messidor-2) confirms the model learns identity-stable anatomical signals rather than dataset-specific shortcuts—the critical property for cross-site deployment. The node insertion/deletion tests further confirm that GoD's attributions are \emph{causally faithful}: high-attribution nodes are the ones the model genuinely relies on, not spurious high-gradient locations. This enables a qualitatively different clinical audit—a radiologist can verify that a match was driven by \emph{right lung boundary and cardiac contour nodes}, rather than interpreting an unstable pixel heatmap. The same node-level attribution directly identifies which anatomical structures carry identity signal, enabling targeted de-identification that suppresses only those structures rather than degrading the whole image.

\noindent\textbf{Limitations.} GoD depends on modality-specific parsers, though our stress test (Table~\ref{tab:robustness}) shows graceful degradation, staying above the no-graph baseline through 50\% node dropout and 10-px mask dilation. It is scoped to 2D, and node attributions are audited at the graph branch rather than propagated through the fusion into $z(x)$. Establishing clinical utility, beyond the faithfulness evidence here, would also require radiologist reader studies.

\section{Conclusion}
We presented Graph-of-Differences (GoD), which grounds medical image re-identification in named anatomy by computing differences over soft-corresponded nodes and aligning them with the global backbone signal. GoD improves retrieval over frozen and fine-tuned baselines, generalises to zero-shot external transfers, and yields node-level explanations with quantitative faithfulness evidence. Future work includes 3D graph correspondence for volumetric CT/MRI, propagating attributions through the fusion into $z(x)$, and a shared anatomical ontology for cross-modality retrieval and attribution-guided de-identification.

\bibliographystyle{splncs04}
\bibliography{mybibliography}

\end{document}